\pdfoutput=1
\documentclass[11pt]{article}

\usepackage{naacl2021}

\usepackage{times}
\usepackage{latexsym}

\usepackage[T1]{fontenc}
\usepackage[utf8]{inputenc}

\usepackage{microtype}
\usepackage{amsfonts}
\usepackage{amsmath}
\usepackage{algorithm, algpseudocode}
\usepackage{graphicx}

\usepackage{verbatim}
\usepackage{stmaryrd}
\usepackage{trimclip}
\usepackage{subcaption}
\usepackage{url}

\makeatletter
\DeclareRobustCommand{\shortto}{%
  \mathrel{\mathpalette\short@to\relax}%
}

\newcommand{\short@to}[2]{%
  \mkern2mu
  \clipbox{{.5\width} 0 0 0}{$\m@th#1\vphantom{+}{\shortrightarrow}$}%
  }
\makeatother
\title{Continual Learning for Text Classification with Information Disentanglement Based Regularization}
\author{Yufan Huang\thanks{\ \ Equal contribution.}, Yanzhe Zhang$^{*1}$, Jiaao Chen, Xuezhi Wang$^2$, Diyi Yang \\
   Georgia Institute of Technology, $^1$Zhejiang University, $^2$Google\\
  \texttt{\{yhuang704, jiaaochen, dyang888\}@gatech.edu}\\ \texttt{$^1$z\_yanzhe@zju.edu.cn, $^2$xuezhiw@google.com}
  }

\begin{document}
\maketitle
\begin{abstract}
Continual learning has become increasingly important as it enables NLP models to constantly learn and gain knowledge over time.  Previous continual learning methods are mainly designed to preserve knowledge from previous tasks, without much emphasis on how to well generalize models to new tasks. In this work, we propose an \textit{information disentanglement} based regularization method for continual learning on text classification. Our proposed method first disentangles text hidden spaces into representations that are generic to all tasks and representations specific to each individual task, and further regularizes these representations differently to better constrain the knowledge required to generalize.  We also introduce two simple auxiliary tasks: next sentence prediction and task-id prediction, for learning better generic and specific representation spaces. Experiments conducted on large-scale benchmarks demonstrate the effectiveness of our method in continual text classification tasks with various sequences and lengths over state-of-the-art baselines. We have publicly released our code at \url{https://github.com/GT-SALT/IDBR}.
\end{abstract}

\section{Introduction}
Computational systems in real world scenarios face changing environment frequently, and thus are often required to learn continually from dynamic streams of data building on what was learnt before \cite{biesialska-etal-2020-continual}. 
For example, a tweeter classifier needs to deal with trending topics which are constantly emerging.
While being an intrinsic nature of human to continually acquire and transfer knowledge throughout lifespans, most machine learning models often suffer from \textit{catastrophic forgetting}: when learning on new tasks, models dramatically and rapidly forget knowledge from previous tasks \cite{mccloskey1989catastrophic}.
As a result, Continual Learning (CL) \citep{ring1998child, thrun1998lifelong} has received more attention recently as it can enable models to perform positive transfer \cite{perkins1992transfer} as well as remember previously seen tasks. 

A growing body of research has been conducted to equip neural networks with the ability of continual learning abilities  \citep{serra2018overcoming,lopez2017gradient,aljundi2018memory}.  % especially in Computer Vision (CV) field, 
% such as applying gradient constraints \citep{kirkpatrick2017overcoming} or regularizing important parameters to prevent the model from forgetting \citep{,lopez2017gradient,aljundi2018memory} on image classification tasks. 
Existing continual learning methods on NLP tasks can be broadly categorized into two classes: purely \emph{replay} based methods \cite{d2019episodic,sun2019lamol} where examples from previous tasks are stored and re-trained during the learning of the new task to retain old information, and \emph{regularization} based methods \cite{wang2019sentence, han2020continual} where constraints are added on model parameters to prevent them from changing too much while learning new tasks. 
The former usually stores an extensive amount of data from old tasks \cite{d2019episodic} or trains language models based on task identifiers to generate sufficient examples \cite{sun2019lamol}, which significantly increases memory costs and training time. 
While the latter utilizes previous examples efficiently  
via the constraints added on text hidden space or model parameters,
it generally views them as equally important and regularize them to the same extent \cite{wang2019sentence, han2020continual}, making it hard for models to differentiate informative representation that needs to be retained from ones that need a large degree of updates. 
However, we argue that when learning new tasks, 
task generic information and task specific information should be treated differently, as these generic representation might function consistently while task specific representations might need to be changed significantly. 

To this end, we propose an information disentanglement based regularization method for continual learning on text classification. Specifically, we first disentangle the text hidden representation space (e.g., the output representation of BERT \citep{devlin-etal-2019-bert}) into a \textbf{task generic} space and a \textbf{task specific} space using two auxiliary tasks:  
next sentence prediction for learning task generic information and task identifier prediction for learning task specific representations. When training on new tasks, 
we constrain the task generic representation to be relatively stable and representations of task specific aspects to be more flexible. 
To further alleviate catastrophic forgetting without much increases of memory and training time, we propose to augment our regularization-based methods by storing and replaying only a small amount of representative examples (e.g., 1\% samples selected by memory selection rules like K-Means
\cite{macqueen1967some}).
To sum up, our contributions are threefold: 
\begin{itemize}
    \item We propose an information disentanglement based regularization method for continual text classification, to better learn and constrain task generic and task specific knowledge. 
    \item We augment the regularization approach with a memory selection rule that requires only a small amount of replaying examples.
    \item Extensive experiments conducted on five benchmark datasets demonstrate the effectiveness of our proposed methods compared to state-of-the-art baselines. 
\end{itemize}

\section{Related work}

\paragraph{Continual Learning} Existing continual learning research can be broadly divided into four categories: (i) replay-based method, which remind models of information from seen tasks via experience replay \citep{d2019episodic}, distillation \citep{rebuffi2017icarl}, representation alignment \citep{wang2019sentence} or optimization constraints \citep{lopez2017gradient, chaudhry2018efficient} using examples sampled from previous tasks \citep{rebuffi2017icarl,d2019episodic} or synthesized with generative models \citep{shin2017continual, sun2019lamol}; (ii) regularization-based method, which constrains model's output \citep{li2017learning}, hidden space \citep{rannen2017encoder}, or parameters \citep{lopez2017gradient, zenke2017continual, aljundi2018memory} from changing too much to retain learned knowledge; (iii) architecture-based method, where different tasks are associated with different components of the overall model to directly minimize the interference between new tasks and old tasks \citep{rusu2016progressive, mallya2018packnet}; (iv) meta-learning-based method, which directly optimizes the knowledge transfer among tasks \citep{riemer2018learning, obamuyide2019meta}, or learns robust data representations \citep{javed2019meta, holla2020meta, wang2020efficient} to alleviate forgetting. 

Among these different approaches, replay-based methods and regularization-based methods have been widely applied to NLP tasks to enable large pre-trained models \cite{devlin-etal-2019-bert,radford2019language} to continually acquire novel world knowledge from streams of textual data without forgetting the already learned knowledge. For instance, replaying examples have shown promising performance for text classification \cite{d2019episodic, sun2019lamol, holla2020meta}, relation extraction \cite{wang2019sentence} and question answering \cite{d2019episodic, sun2019lamol, wang2020efficient}. However, they often suffer from large memory costs or considerable training time, due to the requirements of storing an extensive amount of texts \cite{d2019episodic} or training language models to generate a sufficient number of examples \cite{sun2019lamol}. Recently, regularization-based methods \cite{wang2019sentence, han2020continual} have also been applied to directly constrain knowledge deposited in model parameters without abundant rehearsal examples. Despite better efficiency compared to replay-based methods, current regularization-based approaches often fail to generalize well to new tasks as they treat and constrain all the information \textit{equally} and thus limit the needed updates for parameters that are specific to different tasks. To overcome these limitations, we propose to first distinguish hidden spaces that need to be retained from those that need to be updated substantially through information disentanglement, and then regularize different spaces \textit{separately}, to better remember previous knowledge as well as transfer to new tasks. In addition, we enhance our regularization method by replaying only a limited amount of examples selected by K-means as the memory selection rule.

\paragraph{Textual Information Disentanglement} Our work is related to information disentanglement for text data, which has been extensively explored in generation tasks like style transfer \cite{fu2017style, zhao2018adversarially, romanov2018adversarial, li2020complementary}, where text hidden representations are often disentangled into sentiment \citep{fu2017style, john2018disentangled}, content \citep{romanov2018adversarial, bao-etal-2019-generating} and syntax \citep{bao-etal-2019-generating} information through supervised learning from pre-defined labels \cite{john2018disentangled} or unsupervised learning with adversarial training \cite{fu2017style,li2020complementary}. Building on these prior works, we differentiate task generic space from task specific space via supervision from two simple yet effective auxiliary tasks: next sentence prediction and task identifier prediction.
% achieve the 
%Textual information disentanglement has been extensively explored in language generation and style transfer. Researchers try to disentangle certain features from neural network's hidden space then they can control certain attributes in the hidden space as they want. In order to make sure certain information is captured by one latent space, a common approach is to directly build a predictor on it, which could be designed to predict (i) sentiment labels, such as positive and negative \citep{fu2017style, john2018disentangled}. (ii) semantic information, such as bag of words distribution \citep{romanov2018adversarial, bao-etal-2019-generating}. (iii) syntactic information, such as constituency parse tree \citep{bao-etal-2019-generating}. Besides predictor, a similar discriminator could be added on the other hidden space to make sure certain information is not captured \citep{fu2017style, zhao2018adversarially, romanov2018adversarial, li2020complementary}. In our approach, we only add two predictor on two separate hidden space to make one task-general and the other one task-specific, which is demonstrated effective enough to facilitate continual learning.

\paragraph{Related Learning Paradigms} There exists some other learning paradigms also dealing with multiple tasks, such as multi-task learning \citep{yu2020gradient} and transfer learning \citep{houlsby2019parameter, pfeiffer2020adapterfusion}. However, neither can fit in the scenario of learning multiple tasks sequentially. The former could be adapted to dynamic environments by storing all seen training data and retraining the model after the arrival of new tasks, which highly decreases efficiency and is impractical in deployment. The latter only focuses on the target tasks and ignores catastrophic forgetting on the source tasks. A more thorough discussion can be found in \citet{biesialska-etal-2020-continual}.

\begin{figure*}[h]
\begin{center}
\includegraphics[width=1.0\linewidth]{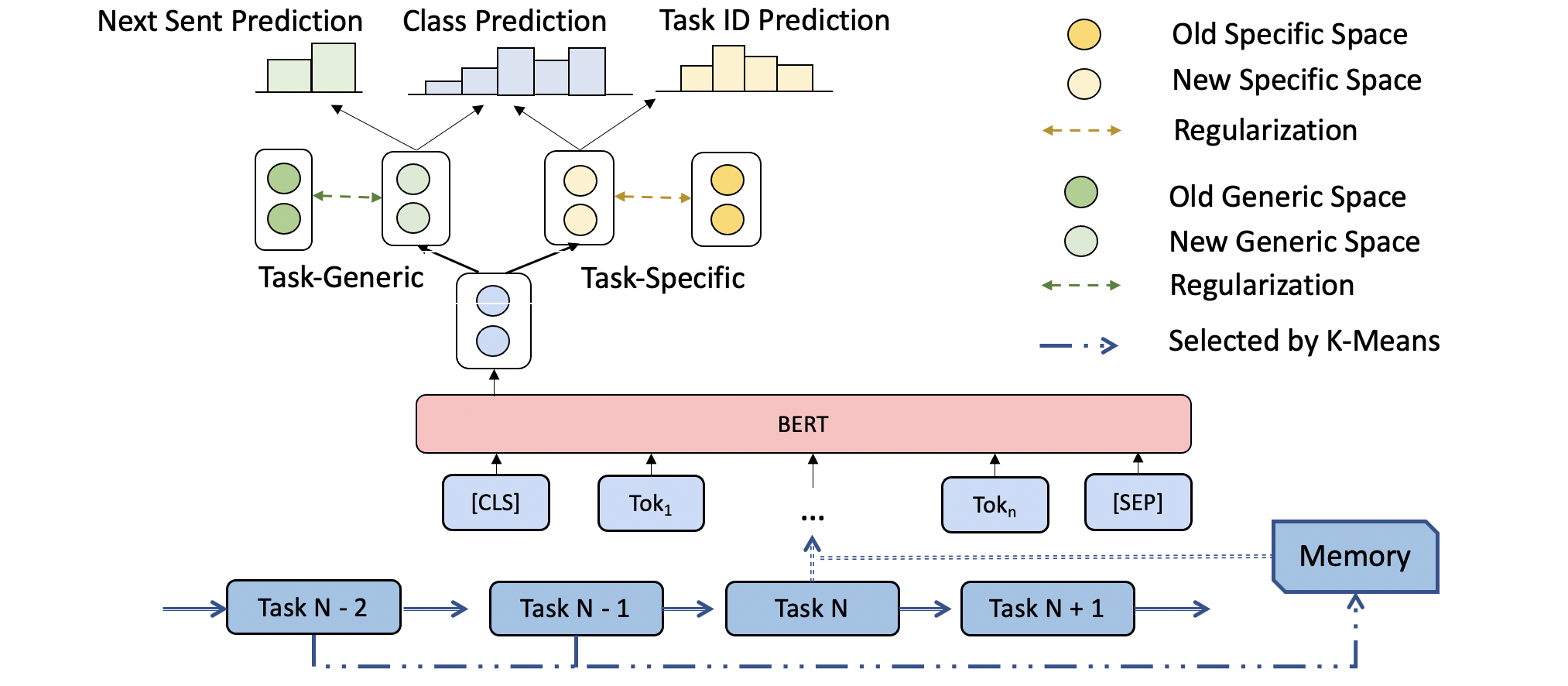}
\end{center}
\caption{Our proposed model architecture. We disentangle the hidden representation into a task generic space and a task specific space via different induction biases. When training on new tasks, different spaces are regularized separately. Also, a small portion of previous data is stored and replayed. %\diyi{the disentaglement is not very explictly visualized here?}
}
\label{fig:model}
\end{figure*}

\section{Problem Formulation}
\label{problem_formulation}
In this work, we focus on continual learning for a sequence of text classification tasks $\{T_1, ... T_n\}$, where we learn a model ${f_\theta(.)}$, $\theta$ is a set of parameters shared by all tasks and each task $T_i$ contains a different set of sentence-label training pairs, $(x^i_{1:m},y^i_{1:m})$. After learning all tasks in the sequence, we seek to minimize the generalization error on all tasks \citep{biesialska-etal-2020-continual}
%\jiaao{Add citations here}
:
\begin{equation*}
    \begin{split}
        R(f_\theta) = \sum_{i=1}^{n} \mathbb{E}_{(x^i, y^i) \sim T_i} \mathcal{L}(f_\theta(x^i), y^i) 
    \end{split}
\end{equation*}
%\yufan{changed $N$ to $n$, and $T^i$ to $T_i$}
We use two commonly-used techniques for this problem setting in our proposed model:
\begin{itemize}
    \item \textbf{Regularization}: in order to preserve knowledge stored in the model, regularization is a constraint added to model output \citep{li2017learning}, hidden space \citep{zenke2017continual} and parameters \citep{lopez2017gradient, zenke2017continual, aljundi2018memory} to prevent them from changing too much while learning new tasks. 
    \item \textbf{Replay}: when learning new tasks, Experience Replay \citep{rebuffi2017icarl} is commonly used to recover knowledge from previous tasks, where a memory buffer is first adopted to store seen examples from previous tasks and then the stored data is replayed with the training set for the current task. Formally, after training on task $t-1\ (t \geq 2)$, $\gamma |S_{t-1}|$ examples are \textit{randomly sampled} from the training set $S_{t-1}$ into the memory buffer $\mathcal{M}$,  where $0\leq \gamma \leq 1$ is the store ratio. Data from $\mathcal{M}$ is then merged with the next training set $S_{t}$ when learning from task $t$.
\end{itemize}

\section{Method}
\label{our-method}
In continual learning, the model needs to adapt to new tasks quickly while maintaining the ability to recover information from previous tasks, hence not all information stored in the hidden representation space should be treated equally.
In previous work like style transfer \cite{john2018disentangled} and controlled text generation \cite{hu2017toward}, certain information (such as content and syntax) is extracted and shared among different categories and other information (such as style and polarity) is manipulated for each specific category. Similarly, in our continual learning scenario, there is shared knowledge among different tasks as well while the model needs to learn and maintain specific knowledge for each individual task in the learning process.
% For example, syntactic knowledge might be shared globally across all tasks, like the ability to recognize word order and grammar; while some knowledge is unique to certain tasks, and should be preserved separately.
This key observation motivates us to propose an \textit{information-disentanglement}  based regularization for continual text classification to retain shared knowledge while adapting specific knowledge to streams of tasks (Section~\ref{Sec: ID}). We also incorporate a small set of representative replay samples to  alleviate catastrophic forgetting (Section~\ref{Sec: Replay}). Our model architecture is shown in Figure \ref{fig:model}. %Formally, for a given sentence $x$, we first use a multi-layer encoder $B(.)$, e.g., BERT \cite{devlin-etal-2019-bert}, to get the hidden representation $r$ which contain both task-general and task-specific information. Then we introduce disentanglement networks $G(.)$ and $S(.)$ to extract the general representation $g$ and specific representation $s$ from $r$ separately.

%The architecture of our method is illustrated in figure \ref{fig:model}. After using a feature extractor $B$ to transform raw input $x$ to embedding $r$, we build two encoders upon $r$, one is task-general encoder $G$, the other is task-specific encoder $S$. By using different induction biases on general representation $g$ and specific representation $s$, we urge $G$ to extract generic information of all tasks, $S$ to extract specific information for each task.

\subsection{Information Disentanglement (ID)} \label{Sec: ID}
This section describes how to disentangle sentence representations into task generic space and task specific space, and how separate regularizations are imposed on them for continual text classification.  Formally, for a given sentence $x$, we first use a multi-layer encoder $B(.)$, e.g., BERT \cite{devlin-etal-2019-bert}, to get the hidden representations $r$ which contain both task generic and task specific information. Then we introduce two disentanglement networks $G(.)$ and $S(.)$ to extract the generic representation $g$ and specific representation $s$ from $r$. For new tasks, we learn the classifiers by utilizing information from both spaces, and we allow different spaces to change to different extents to best retain knowledge from previous tasks.

\paragraph{Task Generic Space}
%\jiaao{What task-general information did we learn? or Could be discussed in experiment section}
%\jiaao{Could visualize the spaces here,or Could be discussed in experiment section}
Task generic space is the hidden space containing information generic to different tasks in a task sequence. During switching from one task to another, the generic information should roughly remain the same, e.g., syntactic knowledge should not change too much across the learning process of a sequence of tasks. 
To extract task generic information $g$ from hidden representations $r$, we leverage the \textit{next sentence prediction} task \citep{devlin-etal-2019-bert} \footnote{Note that the word "sentence" here refers to an arbitrary span of continuous text \citep{devlin-etal-2019-bert}, it could be several linguistic sentences or part of a linguistic sentence.} to learn the generic information extractor $G(.)$.
%create the generic representation $g$. 
More specifically, we insert a \texttt{[SEP]} token into each training example during tokenization to form a sequence pair labeled \emph{IsNext}, and switch the first sequence and the second sequence to form a sentence pair labeled \emph{NotNext}.
% Intuitively, the next sentence prediction task can better help $G(.)$ to capture generic information common to any individual task. 
In order to distinguish \emph{IsNext} pairs and \emph{NotNext} pairs, extractor $G(.)$ needs to learn the context dependencies between two segments, which is beneficial to understand every example and generic to any individual task.
% This is because when doing next sentence prediction,  Those information specific to one task doesn't help in next sentence prediction.
% \jiaao{Please explain why NSP could extract generic information. and define what is generic information!}

Denote $\tilde{x}$ as the \emph{NotNext} example corresponding to $x$ (\emph{IsNext}), and $l\in\{0, 1\}$ as the label for next sentence prediction. We build a sentence relation predictor $f_{nsp}$ on the generic feature extractor $G(.)$: 
\begin{equation*}
    \begin{split}
        \mathcal{L}_{nsp} = \mathbb{E}_{x \in S_t \cup \mathcal{M}} ( &\mathcal{L}(f_{nsp}(G(B(x)), 0) \\
         + &\mathcal{L}(f_{nsp}(G(B(\tilde{x})), 1))
    \end{split}
\end{equation*}
where $\mathcal{L}$ is the cross entropy loss, $\mathcal{M}$ is the memory buffer and $S_{t}$ is the $t$-th training set.

\paragraph{Task Specific Space} %\jiaao{What task-specific information did we learn? or Could be discussed in experiment section}
%\jiaao{Could visualize the spaces here, or Could be discussed in experiment section}
% \jiaao{Likewise, please elaborate what is task-specific information! And why task id prediction could extract those information!}
% In addition to the task generic space, 
Models also need task specific information to perform well over each task. For example, on sentiment classification words like ``good'' or ``bad'' could be very informative, but they might not generalize well for tasks like topic classification.
Thus we employ a simple task-identifier prediction task on the task specific representation $s$, which means for any given example we want to distinguish which task this example belongs to. This simple auxiliary setup will encourage $s$ to embed different information from different tasks. The loss for task-identifier predictor $f_{task}$ is:
\begin{equation*}
    \begin{split}
        \mathcal{L}_{task} = \mathbb{E}_{(x, z) \in S_t \cup \mathcal{M}} \mathcal{L}(f_{task}(S(B(x)), z) 
    \end{split}
\end{equation*}
where $z$ is the corresponding task id for $x$. 
% where $\mathcal{L}$ is the cross entropy loss, $z$ is the corresponding task id for $x$, $\mathcal{M}$ is the memory buffer and $S_{t}$ is the $t$-th training set.
% \xuezhi{to confirm, is y used in the equation? what is z? \{tsk\} is not intuitive, can we use \{task\} or \{specific\}?}

% \jiaao{How to you avoid the overlap between different spaces?}

\paragraph{Text Classification} To adapt to the $t$-th task, we combine the task generic representation $g = G(B(x))$ and task specific representation $s = S(B(x))$ to perform text classification, where we minimize the cross entropy loss: 
% \paragraph{Text classification} 
%Finally, we combine the disentangled task-general representation $g$ and task-specific representation $s$ for text classification. The loss term for classification is:
\begin{equation*}
    \begin{split}
        \mathcal{L}_{cls} = \mathbb{E}_{(x, y) \in S_t \cup \mathcal{M}} \mathcal{L}(f_{cls}(g \circ s), y)) 
    \end{split}
\end{equation*}
Here $y$ is the corresponding class label for $x$, $f_{cls}(.)$ is the class predictor. % , $\mathcal{M}$ is the memory buffer and $S_{t}$ is the $t$-th training set. 
$\circ$ denotes the concatenation of the two representations. 

\subsection{ID Based Regularization}
% \jiaao{add descriptions to tell the intuitions, why we need different weight on different space}
% In the meanwhile, 
To further prevent severe distortion when training on new tasks, we employ regularization on both generic representations $g$ and specific representations $s$.
% To mitigate forgetting, 
Different from previous approaches \cite{li2017learning, wang2019sentence} which treat all the spaces equally, we allow regularization to different extents on $g$ and $s$ as knowledge in different spaces should be preserved separately to encourage both more positive transfer and less forgetting. 
Specifically, before training all the modules on task $t$, we first compute the generic representations and specific representations of all sentences $x$ from the training set $S_t$ of current task $t$ and memory buffer $\mathcal{M}_t$.
Using the trained $B^{t-1}(.)$, $G^{t-1}(.)$ and $S^{t-1}(.)$ from previous task $t-1$, for each example $x$ we calculate the generic representation as $G^{t-1}(B^{t-1}(x))$, and the specific representation as $S^{t-1}(B^{t-1}(x))$ to hoard the knowledge from previous models. The computed generic and specific representations are saved. During the learning from training pairs from task $t$, we impose two regularization losses separately:
\begin{align*}
    \begin{split}
        & \mathcal{L}_{reg}^{g} = \mathbb{E}_{x \in S_t \cup \mathcal{M}_t} \| G^{t-1}(B^{t-1}(x)) - G(B(x)) \|_2 \\
        & \mathcal{L}_{reg}^{s} = \mathbb{E}_{x \in S_t \cup \mathcal{M}_t} \| S^{t-1}(B^{t-1}(x)) - S(B(x)) \|_2
    \end{split}
\end{align*}
% where $\mathcal{M}$ is the memory buffer and $S_{t}$ is the $t$-th training set.
%\xuezhi{i think a better way to write this is using t, t+1 as superscripts}

\subsection{Memory Selection Rule} \label{Sec: Replay}
% \diyi{this is not mentioned in the intro, make sure names are consistent}
Since we only store a small number of examples as a way to balance the replay as well as the extra memory cost and training time, we need to carefully select them in order to utilize the memory  buffer $\mathcal{M}$ efficiently. Considering that if two stored examples are very similar, then only storing one of them could possibly achieve similar results in the future. Thus, those stored examples should be as diverse and representative as possible. To this end, after training on $t$-th task, we employ K-means \citep{macqueen1967some} to cluster all the examples from current training set $S_{t}$: For each $x \in S_{t}$, we utilize its embedding $B(x)$ as its input feature to conduct K-means. We set the numbers of clusters to $\gamma |S_t|$ and only select the example closest to each cluster's centroid, following \citet{wang2019sentence, han2020continual}.

\subsection{Overall Objective}
% \jiaao{aggregate different objectives with weights}
% \xuezhi{the objective might raise some questions, reviewers might ask why so many parameters and how to practically tune them? can we make this more elegant? maybe separate nsp/tsk as representation learning and g/s as regularization}
We can write the final objective for continual learning on text classification as the following:
\begin{equation}
\label{loss}
    \begin{split}
    \mathcal{L} = \mathcal{L}_{cls} + \mathcal{L}_{nsp} + \mathcal{L}_{task} \\ + \lambda^{g} \mathcal{L}_{reg}^{g} + \lambda^{s} \mathcal{L}_{reg}^{s}
    \end{split}
\end{equation}
We set the coefficient of the first three loss terms to 1 for simplicity and only introduce two coefficients to tune: $\lambda^{g}$ and $\lambda^{s}$. In practice, $\mathcal{L}_{task}$ and $\mathcal{L}_{cls}$ are also conducted on each generated \emph{NotNext} example $\tilde{x}$, $\mathcal{L}_{reg}^{g}$ and $\mathcal{L}_{reg}^{s}$ are only optimized starting from the second task. The full information disentanglement based regularization (IDBR) algorithm is shown in Algorithm~\ref{alg_idbr}.

\begin{algorithm}[ht!]
    \caption{IDBR}
    \label{pesudo-code}
    \textbf{Input} Training sets $\{S_1,..., S_n\}$, Replay Frequency $\beta$, Store ratio $\gamma$, Coefficients $\lambda_{g}, \lambda_{s}$\\
    \textbf{Output} Optimal models $B$, $G$, $S$, $f_{nsp}$, $f_{task}$, $f_{cls}$
    \begin{algorithmic} % The number tells where the line numbering should start
            \State $\mathcal{M} = \{\}$  \Comment{Initialize memory buffer} 
            \State Initialize $B$ using pretrained BERT
            \State Initialize $G, S, f_{nsp}, f_{task}, f_{cls}$
            \For{$t = 1,\ldots, n$}
                \If{$t \geq 2$}
                    \State Store $G(B(x))$, $S(B(x))$, $\forall x \in S_t \cup \mathcal{M}$
                    \For{ batches $\in S_t$}
                        % \State Create \emph{NotNext} examples
                        \State Optimize $\mathcal{L}$ in Equation \ref{loss}
                        \If{step $\mod \beta = 0$}
                            \Comment{Replay}
                            \State Sample $t - 1$ batches from $\mathcal{M}$
                            % \State Create \emph{NotNext} examples
                            \State Optimize $\mathcal{L}$ in Equation \ref{loss}
                        \EndIf
                    \EndFor
                \Else \Comment{No regularization on 1st task}
                    \For{ batches $\in S_t$}
                        % \State Create \emph{NotNext} examples
                        \State Optimize $\mathcal{L} = \mathcal{L}_{cls} + \mathcal{L}_{nsp} + \mathcal{L}_{task}$
                    \EndFor
                \EndIf
                \State $\mathcal{C}$ = K-Means($S_t$, $n_{clusters}$=$\gamma |S_t|$) \Comment{$\mathcal{C}$ : centroid}
                \State $\mathcal{C^{\prime}}$ = \{ Examples closest to centers $\in C$ \}
                \State $\mathcal{M} \leftarrow \mathcal{M} \cup \mathcal{C^{\prime}}$ \Comment{Add to memory}
            \EndFor
            \State \textbf{return} $B$, $G$, $S$, $f_{nsp}$, $f_{task}$, $f_{cls}$ %\Comment{Return the optimal model}
    \end{algorithmic}
    \label{alg_idbr}
\end{algorithm}

\section{Experiment}

\subsection{Datasets}
%\jiaao{Use a table to summarize the datasets we are using: citations, domains, class number, train/dev/test set size}
%\xuezhi{need citation for the datasets, maybe a table summarizing info would be more clear}
Following MBPA++ \citep{d2019episodic}, we use five text classification datasets \citep{zhang2015character,chen-etal-2020-mixtext} to evaluate our methods, including AG News (news classification), Yelp (sentiment analysis), DBPedia (Wikipedia article classification), Amazon (sentiment analysis), and Yahoo! Answer (Q\&A classification).
A summary of the datasets is shown in Table~\ref{table-dataset}.
We merge the label space of Amazon and Yelp considering their domain similarity, with 33 classes in total.

\subsection{Experiment Setup}
% Due to the limitation of resources, for most of our experiments, we randomly sample 2000 examples per class for every task and create a reduced dataset.
Due to the limitation of resources, for most of our experiments, we create a reduced dataset by randomly sampling 2000 training examples and 2000 validation examples per class for every task.
See Table \ref{table-dataset} for the train/test size of each dataset. We name this setting Setting (Sampled). We tune all the hyperparameters on the basis of Setting (Sampled). Beyond that, to have a comparison with previous State-of-the-art, we also conduct experiments on the same training set and test set as MbPA++ \cite{d2019episodic} and LAMOL \citep{sun2019lamol}, which contains 115,000 training examples and 7,600 test examples for each task. For every task, we randomly hold out 500 examples per class from training examples for validation purpose. We name the latter Setting (Full). During training, we evaluate our model on validation sets from all seen tasks, following \citet{serra2018overcoming}.

\begin{table}[t]
\centering
\begin{tabular}{lllll}
\hline
\textbf{Dataset}  & \textbf{Class} & \textbf{Type} & \textbf{Train}& \textbf{Test} \\ \hline
AGNews  & 4 & News & 8000 & 7600  \\
Yelp & 5 & Sentiment& 10000 & 7600 \\
Amazon & 5 & Sentiment & 10000 & 7600 \\
DBPedia & 14 & Wikipedia & 28000 & 7600 \\
Yahoo & 10 & Q\&A & 20000 & 7600 \\
\hline 
\end{tabular}
\caption{\label{table-dataset}
Dataset statistics we used for Setting (Sampled). Type means the domain of task classification. Note that the size of the validation set is the same as the size of the training set.
}
\end{table}

Our experiments are mainly conducted on the task sequences shown in Table \ref{table-order}. To minimize the effect of task order and task sequence length on the results, we examine both length-3 task sequences and length-5 task sequences in various orders. The first 3 task sequences are a cyclic shift of ag $\shortto$ yelp $\shortto$ yahoo, which are three classification tasks in different domains (news classification, sentiment analysis, Q\&A classification). The last four length-5 task sequences follows \citet{d2019episodic}.
    
\begin{table}
\centering
\begin{tabular}{ll}
\hline
\textbf{Order} &  \textbf{Task Sequence} \\
\hline
1 & ag $\shortto$ yelp $\shortto$ yahoo \\
2 & yelp $\shortto$ yahoo $\shortto$ ag \\
3 & yahoo $\shortto$ ag $\shortto$ yelp \\
4 & ag $\shortto$ yelp $\shortto$ amazon $\shortto$ yahoo $\shortto$ dbpedia \\
5 & yelp $\shortto$ yahoo $\shortto$ amazon $\shortto$ dbpedia $\shortto$ ag \\
6 & dbpedia $\shortto$ yahoo $\shortto$ ag $\shortto$ amazon $\shortto$ yelp \\
7 & yelp $\shortto$ ag $\shortto$ dbpedia $\shortto$ amazon $\shortto$ yahoo\\  
\hline 
\end{tabular}
\caption{\label{table-order}
Seven random different task sequences used for experiments. The first 6 are used in Setting (Sampled). The last 4 are used in Setting (Full). % \diyi{why these orders, not others?}
}
\end{table}

\subsection{Baselines}
%\jiaao{Add citations for baseline models!}
We compare our proposed model with the following baselines in our experiments:
\begin{itemize}
    \item \textbf{Finetune} \citep{yogatama2019learning}: finetune BERT model sequentially without the episodic memory module and any other loss. %  function.
    \item \textbf{Replay} \citep{wang2019sentence, d2019episodic}: \textbf{Finetune} model augmented with an episodic memory. Replay examples from old tasks while learning new tasks.
    \item \textbf{Regularization}: On top of \textbf{Replay}, with an L2 regularization term added on the hidden state of the classifier following BERT.
    \item \textbf{MBPA++} \citep{d2019episodic}: augment BERT model with an episodic memory module and store \textbf{all} seen examples. MBPA++ performs experience replay at training time, and uses K-nearest neighbors to select examples for local adaptation at test time.
    \item \textbf{LAMOL} \citep{sun2019lamol}: train a language model  that simultaneously learns to solve the tasks and generate training samples, the latter is for generating pseudo samples used in experience replay. Here the text classification is performed in Q\&A formats.  
    \item \textbf{Multi-task Learning} (MTL): The model is trained on all tasks simultaneously, which can be considered as an upper-bound for continual learning methods since it has access to data from all tasks at the same time.
\end{itemize}

\subsection{Implementation Details}
We use pretrained BERT-base-uncased from HuggingFace Transformers \citep{wolf2020transformers} as our base feature extractor. The task generic encoder and task specific encoder are both one linear layer followed by activation function $Tanh$, their output size are both 128 dimensions. The predictors built on encoders are all one linear layer followed by activation function $softmax$. 

All experiments are conducted on NVIDIA RTX 2080 Ti with 11GB memory with the batch size of 8 and the maximum sequence length of 256 (use the first 256 tokens if one's length is beyond that). We use AdamW \citep{loshchilov2017decoupled} as optimizer. For all modules except the task id predictor, we set the learning rate $lr = 3e^{-5}$; for task id predictor, we set its learning rate $lr_{task} = 5e^{-4}$. The weight decay for all parameters are 0.01.

For experience replay, we set the store ratio $\gamma = 0.01$, i.e. we store 1\% of seen examples into the episodic memory module. Besides, we set the replay frequency $\beta = 10$, which means we do experience replay once every ten steps.

For information disentanglement, we mainly tune the coefficients of the regularization loss. For batches from memory buffer $\mathcal{M}$, we set $\lambda^{g}$ to 2.5, select best $\lambda^{s}$ from $\{1.5, 2.0, 2.5\}$. For batches from current training set $S$, we set $\lambda^{g}$ to 0.25, select best $\lambda^{s}$ from $\{0.15, 0.20, 0.25\}$.

\begin{table*}[t]
\centering
\begin{tabular}{lllllllll}
\hline
\textbf{Model} &  \multicolumn{4}{c}{\textbf{Length-3 Task Sequences}} & \multicolumn{4}{c}{\textbf{Length-5 Task Sequences}}\\
\hline
\textbf{Order}  & \textbf{1} & \textbf{2} & \textbf{3} & \textbf{Average} & \textbf{ 4} & \textbf{5} & \textbf{6} & \textbf{Average}\\
Finetune  & 25.79 & 36.56 & 41.01 & 34.45 & 32.37 & 32.22 & 26.44 & 30.34\\
Replay & 69.32 & 70.25 & 71.31 & 70.29 &68.25 & 70.52 & 70.24 & 69.67\\
Regularization & 71.50 & 70.88 & 72.93 & 71.77 & 72.28 & 73.03 & 72.92 & 72.74\\
IDBR & \textbf{71.80} &\textbf{72.72} &  \textbf{73.08} & \textbf{72.53} &  \textbf{72.63}& \textbf{73.72} & \textbf{73.23} & \textbf{73.19} \\
\hline 
MTL & 74.16 & 74.16 & 74.16 & 74.16 & 75.09 & 75.09 & 75.09 & 75.09\\
\hline 
\end{tabular}
\caption{\label{table-res1}
Summary of results on Setting (Sampled) using averaged accuracy after training on the last task. All results are averaged over 3 runs. The $p$-values of paired $t$-test between nine numbers of Regularization and IDBR are 0.018 on Length-3 and 0.009 on Length-5, demonstrating the significant differences.
}
\end{table*}

\begin{table*}[t]
\centering
\begin{tabular}{llllllllll}
\hline
\textbf{Model} & \textbf{TT} & \textbf{TI} & \textbf{LA} & \textbf{PM} & \multicolumn{5}{c}{\textbf{Length-5 Task Sequences}} \\
\hline
\textbf{Order}  &  &  &  & & \textbf{4} & \textbf{5} & \textbf{6} & \textbf{7} & \textbf{Average} \\
MBPA++ $\dagger$ & & & \checkmark & BERT & 70.7 & 70.2 & 70.9 & 70.8 & 70.7 \\
MBPA++ $\dagger \dagger$ & & & \checkmark & BERT & 74.9& 73.1 & 74.9 & 74.1 & 74.3 \\
LAMOL  $\dagger \dagger$& \checkmark & \checkmark &&  GPT-2  &  \textbf{76.1} & 76.1 & \textbf{77.2} & \textbf{76.7} & \textbf{76.5}\\
IDBR & \checkmark & & & BERT & 75.9 & \textbf{76.2} & 76.4 & \textbf{76.7} & 76.3 \\
\hline 
\end{tabular}
\caption{\label{table-res2}
Summary of results on Setting (Full) using averaged accuracy after training on the last task. Our results are averaged over 2 runs. $\dagger$ means we fetch numbers from \citet{d2019episodic}. $\dagger \dagger$ means we fetch numbers from \citet{sun2019lamol}. \textbf{TT}: whether task-id is available during training. \textbf{TI}: whether task-id is available during inference. \textbf{LA}: whether need local adaptation during inference. \textbf{PM}: pretrained models used for continual learning.}
\end{table*}

\section{Results and Discussion}
% \subsection{Results}
We evaluate models after training on all tasks and report their average accuracies on all test sets as our metric. Table \ref{table-res1} summarizes our results in Setting (Sampled). While continual finetuning suffered from severe forgetting, experience replay with 1\% stored examples achieves promising results, which demonstrates the importance of experience replay for continual learning in NLP. Beyond that, simple regularization turns out to be a robust method on the basis of experience replay, which shows consistent improvements on all 6 orders. Our proposed Information Disentanglement Based Regularization (IDBR) further improves regularization consistently under all circumstances.

Table \ref{table-res2} compares IDBR with previous SOTA: MBPA++ and LAMOL 
in Setting (Full). 
Note that although we use the same training/testing data, there is some inherent differences between our settings and previous SOTA methods.
Despite the fact that MBPA++ applies local adaptation when testing, IDBR still outperforms it by an obvious margin. We achieve comparative results with LAMOL, despite that LAMOL requires task identifiers during inference which makes its prediction task easier.

\subsection{Impact of the Lengths of Task Sequences}
Comparing results of length-3 sequences and length-5 sequences in Table \ref{table-res1}, we found that the gap between IDBR and multi-task learning became bigger when the length of task sequence changed from 3 to 5.
To better understand how IDBR gradually forgot, we followed \citet{chaudhry2018riemannian} to measure \textit{forgetting} $\mathcal{F}_{k}$ after trained on task $k$ as follows:
\begin{align*}
    \begin{split}
        & \mathcal{F}_{k} = \mathbb{E}_{j = 1 \dots t - 1} f^{k}_{j}, \\
        & f^{k}_{j} = \max_{l \in \{ 1 \dots k - 1\}} a_{l, j} - a_{k, j}
    \end{split}
\end{align*}
where $a_{l, j}$ is the is the model's accuracy on task $j$ after trained on task $l$. On order 4, 5 and 6, we calculate the forgetting every time after IDBR was trained on a new task and summarize them in Table \ref{table-length}. For continual learning, we hypothesize that the model is prone to suffer from more severe forgetting as the task sequence becomes longer.  We found that although there was some big drop after training on the 3rd task, IDBR maintained stable performance as the length of task sequence increased, especially after training on 4-th and 5-th task, the forgetting increment was relatively small, which demonstrated the robustness of IDBR.

\begin{table}[t]
\centering
\begin{tabular}{lllll}
\hline
\textbf{Order}  & \textbf{4} & \textbf{5} & \textbf{6} & \textbf{Average} \\
\hline 
After 2 tasks & 0.64 & 1.63 & 0.07 & 0.78 \\
After 3 tasks & 3.18 & 2.56 & 1.56 & 2.43 \\
After 4 tasks & 3.60 & 2.17 & 2.20 & 2.66 \\
After 5 tasks & 3.46 & 2.33 & 2.88 & 2.89 \\
\hline 
\end{tabular}
\caption{\label{table-length}
Forgetting measure \cite{chaudhry2018riemannian} calculated every time after finishing training on a new task. All results are averaged over 3 runs.}
\end{table}

\subsection{Visualizing Disentangled Spaces}

\begin{figure}[t]
\captionsetup[subfigure]{justification=centering}
\centering
\begin{subfigure}[b]{0.48\linewidth}
   \includegraphics[width=1.0\linewidth]{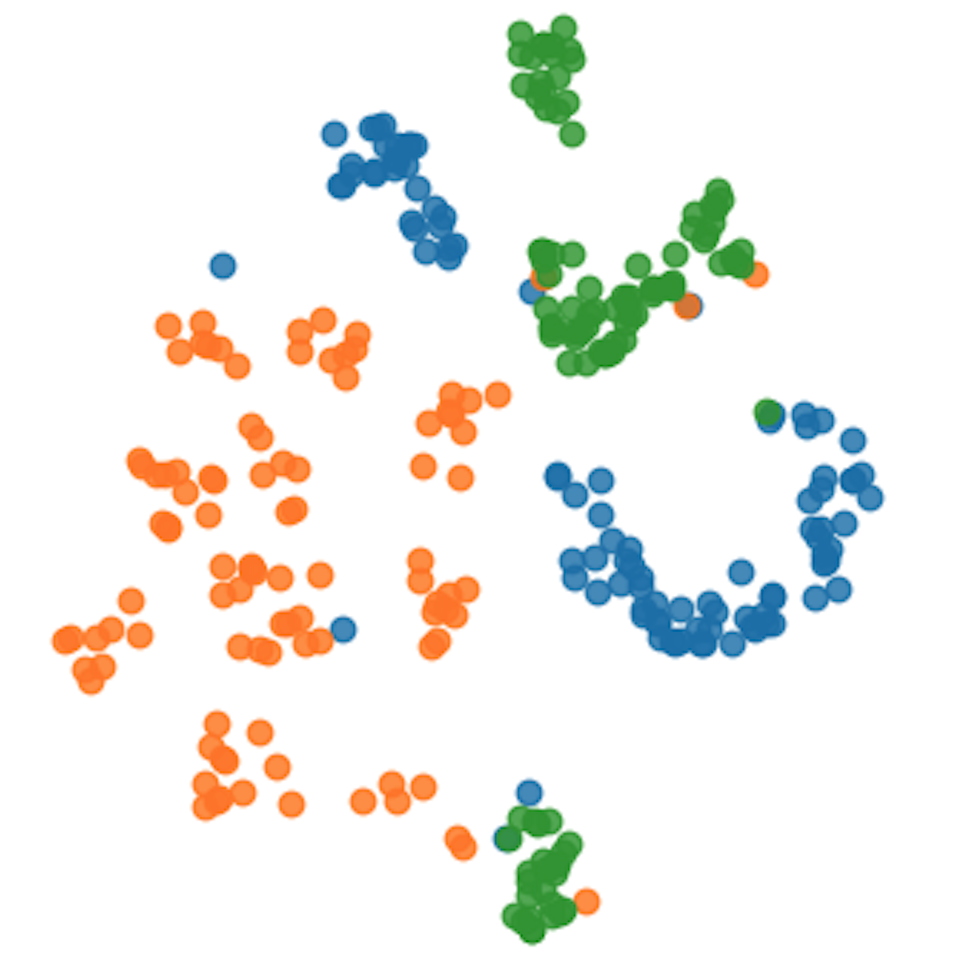}
   \caption{Task Generic Space}
   \label{vis-generic} 
\end{subfigure}
\begin{subfigure}[b]{0.48\linewidth}
   \includegraphics[width=1.0\linewidth]{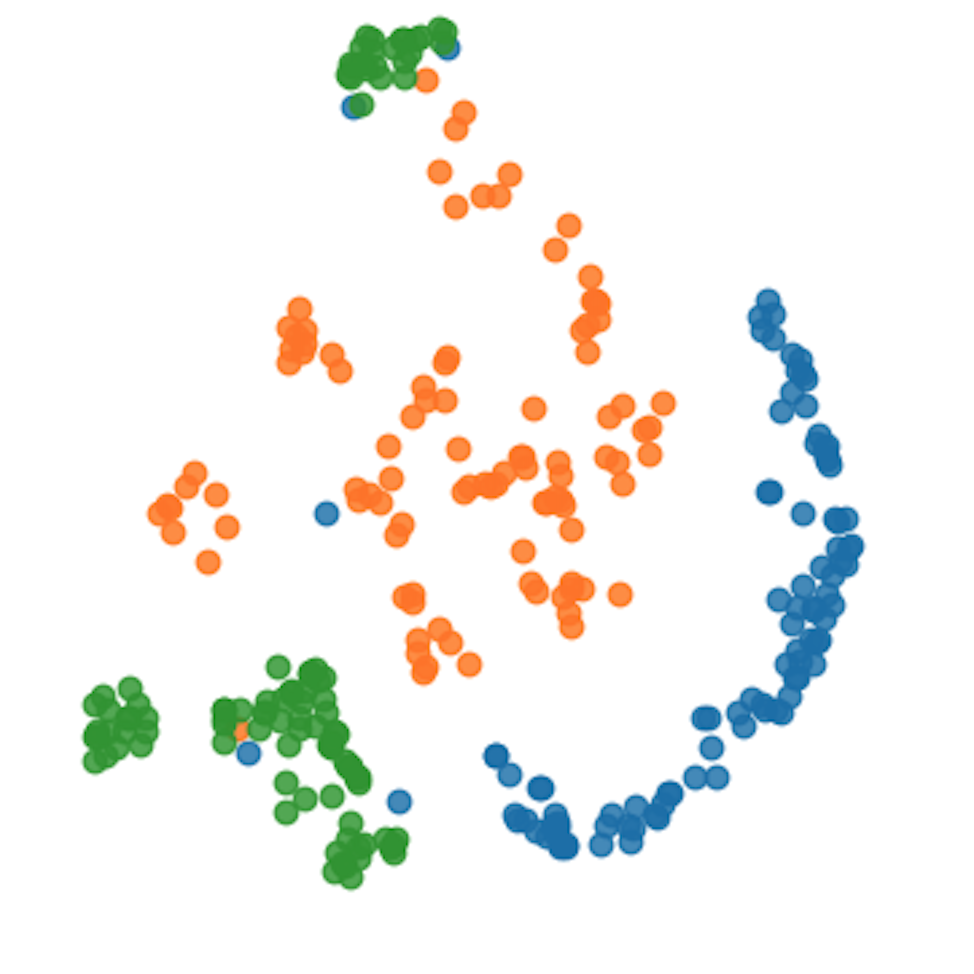}
   \caption{Task Specific Space}
   \label{vis-specific}
\end{subfigure}
\caption{t-SNE visualization of task generic hidden space and task specific hidden space of IDBR.}
\label{vis-hidden}
\end{figure}

To study whether our task generic encoder $G$ tends to learn more generic information and task specific encoder $S$ captures more task specific information, we used t-SNE \cite{maaten2008visualizing} to visualize the two hidden spaces of IDBR, using the final model trained on order 2, and the results are shown in Figure \ref{vis-hidden}, where Figure \ref{vis-generic} visualizes task generic space and Figure \ref{vis-specific} visualizes task specific space. We observe that compared with task specific space, generic features from different tasks were more mixed, which demonstrates that the next sentence prediction helped task generic space to be more task-agnostic than task specific space, which was induced to learn separated representations for different tasks. Considering we only employed two simple auxiliary tasks, the effect of information disentanglement was noticeable. % remarkable. 

\subsection{Ablation Studies}

\begin{table}[t]
\centering
\begin{tabular}{ll}
\hline
\textbf{Model} &  \textbf{Accuracy} \\
\hline
Regularization &  73.03\\
IDBR w/o $\mathcal{L}_{nsp}$& 73.17\\
IDBR w/o $\mathcal{L}_{tsk}$ & 73.29\\
IDBR &  \textbf{73.72}\\ \hline
\end{tabular}
\caption{\label{table-abl}
Comparison among using task-id prediction only, next sentence prediction only and both of them. All results are averaged over 3 runs.
}
\end{table}

\paragraph{Effect of Disentanglement}
In order to demonstrate that each module of our information disentanglement helps the learning process, we performed ablation study on the two auxiliary tasks using order 5 as a case study. The results are summarized in Table \ref{table-abl}. 
We found that both task-id prediction and next sentence prediction contribute to the final performance. Furthermore, the performance gain was much larger by combing these two auxiliary tasks together. Intuitively, the model needs both tasks to disentangle the representation well, since it is easy for the model to ignore one of the spaces if the constraint is not imposed appropriately. The results show that the two tasks are likely complimentary to each other in helping the model learn better disentangled representations.
% The reason was that only adding one auxiliary task made the information disentanglement degenerate to simply adding one auxiliary task to help the encoder learn more information. 
% Only by adding two orthogonal auxiliary tasks on different spaces could help the model learn better disentangled representations.

\begin{table}[t]
\centering
\begin{tabular}{lllll}
\hline
\textbf{Model}  & \textbf{4} & \textbf{5} & \textbf{6} & \textbf{Avg} \\
\hline 
Reg only on $s$  & 72.05  & 72.54 & 72.61 & 72.40 \\
Reg only on $g$ & 72.01 & 72.98 &  72.73 & 72.57\\
Reg on both & \textbf{72.63} & \textbf{73.72} & \textbf{73.23} & \textbf{73.19} \\
\hline 
\end{tabular}
\caption{\label{table-abl-2}
Comparison among using regularization on task specific space only, task generic space only and both of them. All results are averaged over 3 runs.
}
\end{table}

\paragraph{Impact of Regularization}
To study the effect of regularization on task generic hidden space $g$ and task specific hidden space $s$, we performed an ablation study which only applied regularization on $g$ or $s$, and compared the results with regularization on both in Table \ref{table-abl-2}.
We found that regularization on both spaces results in a much better performance than regularization on one of them only, which demonstrates the necessity of both regularizers. While we may expect to give more tolerance to specific space for changing, 
we found that no regularization on it would lead to severe forgetting of previously learnt good task specific embeddings, hence it is necessary to add a regularizer over this space as well. Beyond that, we also observed that under most circumstances, adding regularization on the task generic space $g$ results in a more significant gain than adding regularization on the task specific space $s$, consistent with our intuition that task generic space changes less across tasks and thus preserving it better helps more in alleviating catastrophic forgetting.

\begin{table}[t]
\centering
\begin{tabular}{lllll}
\hline
\textbf{Rules}  & \textbf{1} & \textbf{2} & \textbf{3} & \textbf{Average} \\
\hline 
Random  & 71.52  & 72.60 & 73.03 & 72.38 \\
K-Means & \textbf{71.80} & \textbf{72.72} & \textbf{73.08}& \textbf{72.53}\\
\hline 
\end{tabular}
\caption{\label{table-abl-3}
Comparison between different selection rules: select stored examples randomly or select by K-Means. All results are averaged over 3 runs.
}
\end{table}

\paragraph{Impact of K-Means} To demonstrate our hypothesis that when the memory budget is limited, selecting the most representative subset of examples is vital to the success of continual learning, we performed an ablation study on order 1,2,3 using IDBR with and without K-Means. The result is shown in Table \ref{table-abl-3}.
From the table, we found that using K-Means helps boost the overall performance. Specifically, the improvement brought by K-Means was larger on those challenging orders, i.e. orders on which IDBR had worse performance. This is because for these challenging orders, the forgetting is more severe and the model needs more examples from previous tasks to help it retain previous knowledge. Thus with the same memory budget constraint, diversity across saved examples will help the model better recover knowledge learned from previous tasks.

\section{Conclusion}
In this work, we introduce an information disentanglement based regularization (IDBR) method for continual text classification, where we disentangle the hidden space into task generic space and task specific space and further regularize them differently. We also leverage K-Means as the memory selection rule to help the model benefit from the augmented episodic memory module. 
Experiments conducted on five benchmark datasets demonstrate that IDBR achieves better performances compared to previous state-of-the-art baselines on sequences of text classification tasks with various orders and lengths. % Our ablation studies show the importance of different auxiliary tasks for information disentanglement and the used regularization. % , it also demonstrate the effect of K-Means as a memory selection rule, comparing to random sampling used before.
% \diyi{add a sentence or two for future work}
We believe the proposed approach can be extended to continual learning for other NLP tasks such as sequence generation and sequence labeling as well, and plan to explore them in the future.

\section*{Acknowledgment }
We would like to thank the anonymous reviewers
for their helpful comments, and the members of Georgia Tech SALT group for their feedback. % This work is supported in part by grants from Google, Amazon and Salesforce.

% \vspace{-0.15in}
\bibliography{anthology,custom}

\begin{thebibliography}{45}
\expandafter\ifx\csname natexlab\endcsname\relax\def\natexlab#1{#1}\fi

\bibitem[{Aljundi et~al.(2018)Aljundi, Babiloni, Elhoseiny, Rohrbach, and
  Tuytelaars}]{aljundi2018memory}
Rahaf Aljundi, Francesca Babiloni, Mohamed Elhoseiny, Marcus Rohrbach, and
  Tinne Tuytelaars. 2018.
\newblock Memory aware synapses: Learning what (not) to forget.
\newblock In \emph{Proceedings of the European Conference on Computer Vision
  (ECCV)}, pages 139--154.

\bibitem[{Bao et~al.(2019)Bao, Zhou, Huang, Li, Mou, Vechtomova, Dai, and
  Chen}]{bao-etal-2019-generating}
Yu~Bao, Hao Zhou, Shujian Huang, Lei Li, Lili Mou, Olga Vechtomova, Xin-yu Dai,
  and Jiajun Chen. 2019.
\newblock \href {https://doi.org/10.18653/v1/P19-1602} {Generating sentences
  from disentangled syntactic and semantic spaces}.
\newblock In \emph{Proceedings of the 57th Annual Meeting of the Association
  for Computational Linguistics}, pages 6008--6019, Florence, Italy.
  Association for Computational Linguistics.

\bibitem[{Biesialska et~al.(2020)Biesialska, Biesialska, and
  Costa-juss{\`a}}]{biesialska-etal-2020-continual}
Magdalena Biesialska, Katarzyna Biesialska, and Marta~R. Costa-juss{\`a}. 2020.
\newblock \href {https://doi.org/10.18653/v1/2020.coling-main.574} {Continual
  lifelong learning in natural language processing: A survey}.
\newblock In \emph{Proceedings of the 28th International Conference on
  Computational Linguistics}, pages 6523--6541, Barcelona, Spain (Online).
  International Committee on Computational Linguistics.

\bibitem[{Chaudhry et~al.(2018)Chaudhry, Dokania, Ajanthan, and
  Torr}]{chaudhry2018riemannian}
Arslan Chaudhry, Puneet~K. Dokania, Thalaiyasingam Ajanthan, and Philip H.~S.
  Torr. 2018.
\newblock Riemannian walk for incremental learning: Understanding forgetting
  and intransigence.
\newblock In \emph{Proceedings of the European Conference on Computer Vision
  (ECCV)}.

\bibitem[{Chaudhry et~al.(2019)Chaudhry, Ranzato, Rohrbach, and
  Elhoseiny}]{chaudhry2018efficient}
Arslan Chaudhry, Marc’Aurelio Ranzato, Marcus Rohrbach, and Mohamed
  Elhoseiny. 2019.
\newblock \href {https://openreview.net/forum?id=Hkf2_sC5FX} {Efficient
  lifelong learning with a-{GEM}}.
\newblock In \emph{International Conference on Learning Representations}.

\bibitem[{Chen et~al.(2020)Chen, Yang, and Yang}]{chen-etal-2020-mixtext}
Jiaao Chen, Zichao Yang, and Diyi Yang. 2020.
\newblock \href {https://doi.org/10.18653/v1/2020.acl-main.194} {{M}ix{T}ext:
  Linguistically-informed interpolation of hidden space for semi-supervised
  text classification}.
\newblock In \emph{Proceedings of the 58th Annual Meeting of the Association
  for Computational Linguistics}, pages 2147--2157, Online. Association for
  Computational Linguistics.

\bibitem[{de~Masson~d\textquotesingle Autume
  et~al.(2019)de~Masson~d\textquotesingle Autume, Ruder, Kong, and
  Yogatama}]{d2019episodic}
Cyprien de~Masson~d\textquotesingle Autume, Sebastian Ruder, Lingpeng Kong, and
  Dani Yogatama. 2019.
\newblock \href
  {https://proceedings.neurips.cc/paper/2019/file/f8d2e80c1458ea2501f98a2cafadb397-Paper.pdf}
  {Episodic memory in lifelong language learning}.
\newblock In \emph{Advances in Neural Information Processing Systems},
  volume~32. Curran Associates, Inc.

\bibitem[{Devlin et~al.(2019)Devlin, Chang, Lee, and
  Toutanova}]{devlin-etal-2019-bert}
Jacob Devlin, Ming-Wei Chang, Kenton Lee, and Kristina Toutanova. 2019.
\newblock \href {https://doi.org/10.18653/v1/N19-1423} {{BERT}: Pre-training of
  deep bidirectional transformers for language understanding}.
\newblock In \emph{Proceedings of the 2019 Conference of the North {A}merican
  Chapter of the Association for Computational Linguistics: Human Language
  Technologies, Volume 1 (Long and Short Papers)}, pages 4171--4186,
  Minneapolis, Minnesota. Association for Computational Linguistics.

\bibitem[{Fu et~al.(2017)Fu, Tan, Peng, Zhao, and Yan}]{fu2017style}
Zhenxin Fu, Xiaoye Tan, Nanyun Peng, Dongyan Zhao, and Rui Yan. 2017.
\newblock Style transfer in text: Exploration and evaluation.
\newblock \emph{arXiv preprint arXiv:1711.06861}.

\bibitem[{Han et~al.(2020)Han, Dai, Gao, Lin, Liu, Li, Sun, and
  Zhou}]{han2020continual}
Xu~Han, Yi~Dai, Tianyu Gao, Yankai Lin, Zhiyuan Liu, Peng Li, Maosong Sun, and
  Jie Zhou. 2020.
\newblock \href {https://doi.org/10.18653/v1/2020.acl-main.573} {Continual
  relation learning via episodic memory activation and reconsolidation}.
\newblock In \emph{Proceedings of the 58th Annual Meeting of the Association
  for Computational Linguistics}, pages 6429--6440, Online. Association for
  Computational Linguistics.

\bibitem[{Holla et~al.(2020)Holla, Mishra, Yannakoudakis, and
  Shutova}]{holla2020meta}
Nithin Holla, Pushkar Mishra, Helen Yannakoudakis, and Ekaterina Shutova. 2020.
\newblock \href {https://arxiv.org/abs/2009.04891} {Meta-learning with sparse
  experience replay for lifelong language learning}.
\newblock \emph{CoRR}, abs/2009.04891.

\bibitem[{Houlsby et~al.(2019)Houlsby, Giurgiu, Jastrzebski, Morrone,
  De~Laroussilhe, Gesmundo, Attariyan, and Gelly}]{houlsby2019parameter}
Neil Houlsby, Andrei Giurgiu, Stanislaw Jastrzebski, Bruna Morrone, Quentin
  De~Laroussilhe, Andrea Gesmundo, Mona Attariyan, and Sylvain Gelly. 2019.
\newblock \href {http://proceedings.mlr.press/v97/houlsby19a.html}
  {Parameter-efficient transfer learning for {NLP}}.
\newblock In \emph{Proceedings of the 36th International Conference on Machine
  Learning}, volume~97 of \emph{Proceedings of Machine Learning Research},
  pages 2790--2799. PMLR.

\bibitem[{Hu et~al.(2017)Hu, Yang, Liang, Salakhutdinov, and
  Xing}]{hu2017toward}
Zhiting Hu, Zichao Yang, Xiaodan Liang, Ruslan Salakhutdinov, and Eric~P. Xing.
  2017.
\newblock \href {http://proceedings.mlr.press/v70/hu17e.html} {Toward
  controlled generation of text}.
\newblock In \emph{Proceedings of the 34th International Conference on Machine
  Learning}, volume~70 of \emph{Proceedings of Machine Learning Research},
  pages 1587--1596. PMLR.

\bibitem[{Javed and White(2019)}]{javed2019meta}
Khurram Javed and Martha White. 2019.
\newblock \href
  {https://proceedings.neurips.cc/paper/2019/file/f4dd765c12f2ef67f98f3558c282a9cd-Paper.pdf}
  {Meta-learning representations for continual learning}.
\newblock In \emph{Advances in Neural Information Processing Systems},
  volume~32. Curran Associates, Inc.

\bibitem[{John et~al.(2019)John, Mou, Bahuleyan, and
  Vechtomova}]{john2018disentangled}
Vineet John, Lili Mou, Hareesh Bahuleyan, and Olga Vechtomova. 2019.
\newblock \href {https://doi.org/10.18653/v1/P19-1041} {Disentangled
  representation learning for non-parallel text style transfer}.
\newblock In \emph{Proceedings of the 57th Annual Meeting of the Association
  for Computational Linguistics}, pages 424--434, Florence, Italy. Association
  for Computational Linguistics.

\bibitem[{Kirkpatrick et~al.(2017)Kirkpatrick, Pascanu, Rabinowitz, Veness,
  Desjardins, Rusu, Milan, Quan, Ramalho, Grabska-Barwinska, Hassabis, Clopath,
  Kumaran, and Hadsell}]{serra2018overcoming}
James Kirkpatrick, Razvan Pascanu, Neil Rabinowitz, Joel Veness, Guillaume
  Desjardins, Andrei~A. Rusu, Kieran Milan, John Quan, Tiago Ramalho, Agnieszka
  Grabska-Barwinska, Demis Hassabis, Claudia Clopath, Dharshan Kumaran, and
  Raia Hadsell. 2017.
\newblock \href {https://doi.org/10.1073/pnas.1611835114} {Overcoming
  catastrophic forgetting in neural networks}.
\newblock \emph{Proceedings of the National Academy of Sciences},
  114(13):3521--3526.

\bibitem[{Li et~al.(2020)Li, Li, Zhang, Li, Zheng, Carin, and
  Gao}]{li2020complementary}
Yuan Li, Chunyuan Li, Yizhe Zhang, Xiujun Li, Guoqing Zheng, Lawrence Carin,
  and Jianfeng Gao. 2020.
\newblock \href {https://doi.org/10.1609/aaai.v34i05.6346} {Complementary
  auxiliary classifiers for label-conditional text generation}.
\newblock \emph{Proceedings of the AAAI Conference on Artificial Intelligence},
  34(05):8303--8310.

\bibitem[{Li and Hoiem(2018)}]{li2017learning}
Zhizhong Li and Derek Hoiem. 2018.
\newblock \href {https://doi.org/10.1109/TPAMI.2017.2773081} {Learning without
  forgetting}.
\newblock \emph{IEEE Transactions on Pattern Analysis and Machine
  Intelligence}, 40(12):2935--2947.

\bibitem[{Lopez-Paz and Ranzato(2017)}]{lopez2017gradient}
David Lopez-Paz and Marc\textquotesingle~Aurelio Ranzato. 2017.
\newblock \href
  {https://proceedings.neurips.cc/paper/2017/file/f87522788a2be2d171666752f97ddebb-Paper.pdf}
  {Gradient episodic memory for continual learning}.
\newblock In \emph{Advances in Neural Information Processing Systems},
  volume~30. Curran Associates, Inc.

\bibitem[{Loshchilov and Hutter(2019)}]{loshchilov2017decoupled}
Ilya Loshchilov and Frank Hutter. 2019.
\newblock \href {https://openreview.net/forum?id=Bkg6RiCqY7} {Decoupled weight
  decay regularization}.
\newblock In \emph{International Conference on Learning Representations}.

\bibitem[{MacQueen et~al.(1967)}]{macqueen1967some}
James MacQueen et~al. 1967.
\newblock Some methods for classification and analysis of multivariate
  observations.
\newblock In \emph{Proceedings of the fifth Berkeley symposium on mathematical
  statistics and probability}, volume~1, pages 281--297. Oakland, CA, USA.

\bibitem[{Mallya and Lazebnik(2018)}]{mallya2018packnet}
Arun Mallya and Svetlana Lazebnik. 2018.
\newblock Packnet: Adding multiple tasks to a single network by iterative
  pruning.
\newblock In \emph{Proceedings of the IEEE Conference on Computer Vision and
  Pattern Recognition}, pages 7765--7773.

\bibitem[{McCloskey and Cohen(1989)}]{mccloskey1989catastrophic}
Michael McCloskey and Neal~J Cohen. 1989.
\newblock Catastrophic interference in connectionist networks: The sequential
  learning problem.
\newblock In \emph{Psychology of learning and motivation}, volume~24, pages
  109--165. Elsevier.

\bibitem[{Obamuyide and Vlachos(2019)}]{obamuyide2019meta}
Abiola Obamuyide and Andreas Vlachos. 2019.
\newblock Meta-learning improves lifelong relation extraction.
\newblock In \emph{Proceedings of the 4th Workshop on Representation Learning
  for NLP (RepL4NLP-2019)}, pages 224--229.

\bibitem[{Perkins et~al.(1992)Perkins, Salomon et~al.}]{perkins1992transfer}
David~N Perkins, Gavriel Salomon, et~al. 1992.
\newblock Transfer of learning.
\newblock \emph{International encyclopedia of education}, 2:6452--6457.

\bibitem[{Pfeiffer et~al.(2021)Pfeiffer, Kamath, R{\"u}ckl{\'e}, Cho, and
  Gurevych}]{pfeiffer2020adapterfusion}
Jonas Pfeiffer, Aishwarya Kamath, Andreas R{\"u}ckl{\'e}, Kyunghyun Cho, and
  Iryna Gurevych. 2021.
\newblock \href {https://www.aclweb.org/anthology/2021.eacl-main.39}
  {{A}dapter{F}usion: Non-destructive task composition for transfer learning}.
\newblock In \emph{Proceedings of the 16th Conference of the European Chapter
  of the Association for Computational Linguistics: Main Volume}, pages
  487--503, Online. Association for Computational Linguistics.

\bibitem[{Radford et~al.(2019)Radford, Wu, Child, Luan, Amodei, and
  Sutskever}]{radford2019language}
Alec Radford, Jeffrey Wu, Rewon Child, David Luan, Dario Amodei, and Ilya
  Sutskever. 2019.
\newblock Language models are unsupervised multitask learners.
\newblock \emph{OpenAI blog}, 1(8):9.

\bibitem[{Rannen et~al.(2017)Rannen, Aljundi, Blaschko, and
  Tuytelaars}]{rannen2017encoder}
Amal Rannen, Rahaf Aljundi, Matthew~B Blaschko, and Tinne Tuytelaars. 2017.
\newblock Encoder based lifelong learning.
\newblock In \emph{Proceedings of the IEEE International Conference on Computer
  Vision}, pages 1320--1328.

\bibitem[{Rebuffi et~al.(2017)Rebuffi, Kolesnikov, Sperl, and
  Lampert}]{rebuffi2017icarl}
Sylvestre-Alvise Rebuffi, Alexander Kolesnikov, Georg Sperl, and Christoph~H
  Lampert. 2017.
\newblock icarl: Incremental classifier and representation learning.
\newblock In \emph{Proceedings of the IEEE conference on Computer Vision and
  Pattern Recognition}, pages 2001--2010.

\bibitem[{Riemer et~al.(2019)Riemer, Cases, Ajemian, Liu, Rish, Tu, , and
  Tesauro}]{riemer2018learning}
Matthew Riemer, Ignacio Cases, Robert Ajemian, Miao Liu, Irina Rish, Yuhai Tu,
  , and Gerald Tesauro. 2019.
\newblock \href {https://openreview.net/forum?id=B1gTShAct7} {Learning to learn
  without forgetting by maximizing transfer and minimizing interference}.
\newblock In \emph{International Conference on Learning Representations}.

\bibitem[{Ring(1998)}]{ring1998child}
Mark~B Ring. 1998.
\newblock Child: A first step towards continual learning.
\newblock In \emph{Learning to learn}, pages 261--292. Springer.

\bibitem[{Romanov et~al.(2019)Romanov, Rumshisky, Rogers, and
  Donahue}]{romanov2018adversarial}
Alexey Romanov, Anna Rumshisky, Anna Rogers, and David Donahue. 2019.
\newblock \href {https://doi.org/10.18653/v1/N19-1088} {Adversarial
  decomposition of text representation}.
\newblock In \emph{Proceedings of the 2019 Conference of the North {A}merican
  Chapter of the Association for Computational Linguistics: Human Language
  Technologies, Volume 1 (Long and Short Papers)}, pages 815--825, Minneapolis,
  Minnesota. Association for Computational Linguistics.

\bibitem[{Rusu et~al.(2016)Rusu, Rabinowitz, Desjardins, Soyer, Kirkpatrick,
  Kavukcuoglu, Pascanu, and Hadsell}]{rusu2016progressive}
Andrei~A Rusu, Neil~C Rabinowitz, Guillaume Desjardins, Hubert Soyer, James
  Kirkpatrick, Koray Kavukcuoglu, Razvan Pascanu, and Raia Hadsell. 2016.
\newblock Progressive neural networks.
\newblock \emph{arXiv preprint arXiv:1606.04671}.

\bibitem[{Shin et~al.(2017)Shin, Lee, Kim, and Kim}]{shin2017continual}
Hanul Shin, Jung~Kwon Lee, Jaehong Kim, and Jiwon Kim. 2017.
\newblock Continual learning with deep generative replay.
\newblock In \emph{Advances in Neural Information Processing Systems}, pages
  2990--2999.

\bibitem[{Sun et~al.(2019)Sun, Ho, and Lee}]{sun2019lamol}
Fan-Keng Sun, Cheng-Hao Ho, and Hung-Yi Lee. 2019.
\newblock Lamol: Language modeling for lifelong language learning.
\newblock In \emph{International Conference on Learning Representations}.

\bibitem[{Thrun(1998)}]{thrun1998lifelong}
Sebastian Thrun. 1998.
\newblock Lifelong learning algorithms.
\newblock In \emph{Learning to learn}, pages 181--209. Springer.

\bibitem[{van~der Maaten and Hinton(2008)}]{maaten2008visualizing}
Laurens van~der Maaten and Geoffrey Hinton. 2008.
\newblock \href {http://jmlr.org/papers/v9/vandermaaten08a.html} {Visualizing
  data using t-sne}.
\newblock \emph{Journal of Machine Learning Research}, 9(86):2579--2605.

\bibitem[{Wang et~al.(2019)Wang, Xiong, Yu, Guo, Chang, and
  Wang}]{wang2019sentence}
Hong Wang, Wenhan Xiong, Mo~Yu, Xiaoxiao Guo, Shiyu Chang, and William~Yang
  Wang. 2019.
\newblock \href {https://doi.org/10.18653/v1/N19-1086} {Sentence embedding
  alignment for lifelong relation extraction}.
\newblock In \emph{Proceedings of the 2019 Conference of the North {A}merican
  Chapter of the Association for Computational Linguistics: Human Language
  Technologies, Volume 1 (Long and Short Papers)}, pages 796--806, Minneapolis,
  Minnesota. Association for Computational Linguistics.

\bibitem[{Wang et~al.(2020)Wang, Mehta, Poczos, and
  Carbonell}]{wang2020efficient}
Zirui Wang, Sanket~Vaibhav Mehta, Barnabas Poczos, and Jaime Carbonell. 2020.
\newblock \href {https://doi.org/10.18653/v1/2020.emnlp-main.39} {Efficient
  meta lifelong-learning with limited memory}.
\newblock In \emph{Proceedings of the 2020 Conference on Empirical Methods in
  Natural Language Processing (EMNLP)}, pages 535--548, Online. Association for
  Computational Linguistics.

\bibitem[{Wolf et~al.(2020)Wolf, Debut, Sanh, Chaumond, Delangue, Moi, Cistac,
  Rault, Louf, Funtowicz, Davison, Shleifer, von Platen, Ma, Jernite, Plu, Xu,
  Le~Scao, Gugger, Drame, Lhoest, and Rush}]{wolf2020transformers}
Thomas Wolf, Lysandre Debut, Victor Sanh, Julien Chaumond, Clement Delangue,
  Anthony Moi, Pierric Cistac, Tim Rault, Remi Louf, Morgan Funtowicz, Joe
  Davison, Sam Shleifer, Patrick von Platen, Clara Ma, Yacine Jernite, Julien
  Plu, Canwen Xu, Teven Le~Scao, Sylvain Gugger, Mariama Drame, Quentin Lhoest,
  and Alexander Rush. 2020.
\newblock \href {https://doi.org/10.18653/v1/2020.emnlp-demos.6} {Transformers:
  State-of-the-art natural language processing}.
\newblock In \emph{Proceedings of the 2020 Conference on Empirical Methods in
  Natural Language Processing: System Demonstrations}, pages 38--45, Online.
  Association for Computational Linguistics.

\bibitem[{Yogatama et~al.(2019)Yogatama, d'Autume, Connor, Kocisky,
  Chrzanowski, Kong, Lazaridou, Ling, Yu, Dyer et~al.}]{yogatama2019learning}
Dani Yogatama, Cyprien de~Masson d'Autume, Jerome Connor, Tomas Kocisky, Mike
  Chrzanowski, Lingpeng Kong, Angeliki Lazaridou, Wang Ling, Lei Yu, Chris
  Dyer, et~al. 2019.
\newblock Learning and evaluating general linguistic intelligence.
\newblock \emph{arXiv preprint arXiv:1901.11373}.

\bibitem[{Yu et~al.(2020)Yu, Kumar, Gupta, Levine, Hausman, and
  Finn}]{yu2020gradient}
Tianhe Yu, Saurabh Kumar, Abhishek Gupta, Sergey Levine, Karol Hausman, and
  Chelsea Finn. 2020.
\newblock \href
  {https://proceedings.neurips.cc/paper/2020/file/3fe78a8acf5fda99de95303940a2420c-Paper.pdf}
  {Gradient surgery for multi-task learning}.
\newblock In \emph{Advances in Neural Information Processing Systems},
  volume~33, pages 5824--5836. Curran Associates, Inc.

\bibitem[{Zenke et~al.(2017)Zenke, Poole, and Ganguli}]{zenke2017continual}
Friedemann Zenke, Ben Poole, and Surya Ganguli. 2017.
\newblock Continual learning through synaptic intelligence.
\newblock \emph{Proceedings of machine learning research}, 70:3987.

\bibitem[{Zhang et~al.(2015)Zhang, Zhao, and LeCun}]{zhang2015character}
Xiang Zhang, Junbo Zhao, and Yann LeCun. 2015.
\newblock Character-level convolutional networks for text classification.
\newblock In \emph{Advances in neural information processing systems}, pages
  649--657.

\bibitem[{Zhao et~al.(2018)Zhao, Kim, Zhang, Rush, and
  LeCun}]{zhao2018adversarially}
Junbo Zhao, Yoon Kim, Kelly Zhang, Alexander Rush, and Yann LeCun. 2018.
\newblock Adversarially regularized autoencoders.
\newblock In \emph{International Conference on Machine Learning}, pages
  5902--5911. PMLR.

\end{thebibliography}
\bibliographystyle{acl_natbib}

\end{document}